\documentclass[iicol,pdflatex,sn-mathphys-num]{sn-jnl}

\usepackage{graphicx}%
\usepackage{multirow}%
\usepackage{amsmath,amssymb,amsfonts}%
\usepackage{amsthm}%
\usepackage{mathrsfs}%
\usepackage[title]{appendix}%
\usepackage{xcolor}%
\usepackage{textcomp}%
\usepackage{manyfoot}%
\usepackage{booktabs}%
\usepackage{algorithm}%
\usepackage{algorithmicx}%
\usepackage{algpseudocode}%
\usepackage{listings}%

\begin{document}

\title[Article Title]{A survey on learning models of spiking neural membrane systems and spiking neural networks}


\author[1]{\fnm{Prithwineel } \sur{Paul}}\email{prithwineel.paul@iem.edu.in}

\author*[2]{\fnm{Petr } \sur{Sos\'\i k}}\email{petr.sosik@fpf.slu.cz}
\equalcont{These authors contributed equally to this work.}

\author[2]{\fnm{Lucie } \sur{Ciencialová}}\email{lucie.ciencialova@fpf.slu.cz}
\equalcont{These authors contributed equally to this work.}

\affil[1]{\orgdiv{Department of Computer Science \& Engineering}, \orgname{Institute of Engineering \& Management, University of Engineering \& Management}, \orgaddress{\street{New Town Rd.}, \city{Kolkata}, \postcode{700091}, \country{ India}}}

\affil*[2]{\orgdiv{Institue of Computer Science}, \orgname{Faculty of Philosophy and Science in Opava, Silesian Univerzity in Opava}, \orgaddress{\street{Bezručovo náměstí 1150/13 }, \city{Opava}, \postcode{746 01}, \country{Czech Rebublic}}}



\abstract{Spiking neural networks (SNN) are a biologically inspired model of neural networks with certain brain-like properties. In the past few decades, this model has received increasing attention in computer science community, owing also to the successful phenomenon of deep learning. In SNN,  communication between neurons takes place through the spikes and spike trains. This differentiates these models from the ``standard'' artificial neural networks (ANN) where the frequency of spikes is replaced by real-valued signals. Spiking neural P systems (SNPS) can be considered a branch of SNN based more on the principles of formal automata, with many variants developed within the framework of the membrane computing theory. In this paper, we first briefly compare structure and function,  advantages and drawbacks of SNN and SNPS. A key part of the article is a survey of recent results and applications of machine learning and deep learning models of both SNN and SNPS formalisms. }



\maketitle

\section{Introduction}\label{sec1}

Spiking neural networks (SNN) are a brain-inspired model  of neural communication and computation using individual spikes to transfer information between individual abstract neurons. These models are known as \textit{third-generation neural networks} \cite{maass1997networks,ghosh2009spiking}. Spiking neurons also follow the principle of integrate-and-fire mechanism.  The internal state of the spiking neurons changes with time and when it exceeds a threshold, the postsynaptic neurons spike. Moreover, these neurons encode the information using the timing of the spikes. More specifically, in SNN the encoding of spatio-temporal information is carried by the spikes.
Research into the construction of neuromorphic hardware has gained momentum in the past few years, while it has also promoted the research of SNN. Implementations of SNN promises faster information processing capabilities and also lower energy consumption. However, analog deep learning models are mostly more efficient in real-life applications than SNN which still lack comparably efficient machine learning algorithms as well as programming frameworks. Also, the spiking activity in SNN is discrete and non-differentiable which complicates the implementation of backpropagation-like algorithms based on gradient descent. Therefore, constructing efficient supervised machine learning algorithms for single-layer and multi-layer SNN is a recent research challenge.

Spiking neural P systems (SNPS)  \cite{ionescu2006spiking} are a variant of spiking neural networks introduced by Ionescu, P{\u a}un, Yokomori in 2006. SNPS are also partly inspired by the formal language and automata theory and they belong to the family of computing models called  membrane systems (also P systems) \cite{10.5555/1738939}. Up-to-date information on membrane computing can be found, e.g., in the survey \cite{song2021survey}. Like SNN, the encoding of information in SNPS is based on the timing of spikes of specified neurons, or also by a number of discrete spikes. In the last decade, SNPS models have gained popularity among the  computer science community because of their applicability to many real-life problems \cite{zhang2017real}. Consequently,  many variants of SNPS models have been introduced and their computational capability have been investigated extensively \cite{leporati2022spiking}.

\textit{The motivation behind preparing this survey is as follows: (1) Though there exist survey works on different aspects of SNN\cite{wang2020supervised,auge2021survey,nunes2022spiking,eshraghian2023training} and SNPS\cite{chen2021survey, leporati2022spiking}, no review covering both models has been published yet.
(2) The main issue of both models is the need for efficient learning methods. Therefore, we focused on machine learning and deep learning algorithms for SNN and SNPS. 
(3) The research in both areas accelerated in the last few years and especially new deep learning methods emerged.}

The structure  of the paper provides simultaneously a key to navigate in the cited literature.
Section \ref{section1} lists main types and discusses the structure and function together with their comparison. In Section \ref{section3} we discuss the machine learning and deep learning algorithms based on SNN, and a similar study is provided for SNPS models in Section  \ref{section4}. Finally, Section \ref{section6} resumes and concludes the paper.

\section{Structure and function of SNN and SNPS } \label{section1}

\subsection{Architecture and function of SNN} 

SNN are inspired by the structure and function of biological neurons which communicate using spikes. The neuron spikes only when the membrane potential exceeds a threshold after receiving some spikes from the neighbouring neurons via synaptic connections.  Synapses are associated with adjustable scalar weights whose sign makes them excitatory or inhibitory (increase or decrease the membrane potential). 
The presynaptic task of an SNN is to encode an analogue input into spike trains. Different types of encoding techniques are used, such as rate-based method, temporal coding \cite{auge2021survey,gerstner2014neuronal} etc. Therefore, the membrane potential of the postsynaptic neurons is modulated by the components of the presynaptic neurons.  Hodgkin and Huxley first modelled this phenomenon in 1952 \cite{hodgkin1952quantitative}. Later, many abstract variants of SNN models have been introduced such as spike response model (SRM) \cite{kistler1997reduction}, the Izhikevich neuron model \cite{izhikevich2003simple},  leaky
integrated-and-fire (LIF) neuron \cite{hodgkin1990quantitative,koravuna2023evaluation}, stochastic SNN \cite{morro2017stochastic,rossello2012hardware}, self-organizing SNN \cite{hazan2018unsupervised}, SNN with adaptive structure \cite{wang2014online}, biorealistic SNN \cite{ambroise2013biorealistic}, probabilistic SNN \cite{kasabov2010spike}. 
Among further well-known SNN models we mention 
memristor-based SNN \cite{fouda2020spiking}, brain-inspired SNN \cite{yu2014brain}, including cerebellar-based robotic SNN controller \cite{zahra2022neurorobotic}, SNN with noise-threshold \cite{zhang2017supervised},  stereospike models \cite{ranccon2022stereospike}, xeno-genetic SNN \cite{vellappally2018maintaining}, Izhikevich SNN  \cite{garaffa2021revealing}, NeuCube \cite{kasabov2014neucube}, SpinalFlow \cite{narayanan2020spinalflow},  polychronous SNN \cite{wang2013fpga}, Rmp-SNN \cite{han2020rmp}, Spiking vgg7 \cite{xiang2022spiking}, BioLCNet \cite{ghaemi2023biolcnet}, 
Spiking Capsnet \cite{zhao2022spiking}, attention SNN \cite{yao2022attention}, AutoSNN \cite{na2022autosnn} and more.

In the case of implementation on a special purpose hardware, SNN  models can be used to construct energy-efficient spiking hardware responding highly to event-based sensors\cite{khan2008spinnaker,han2020deep,tang2019spiking}, on one hand. On the other hand, the construction of hardware based on SNPS models is still at a developing stage. 

Industrial applications of SNN lags behind the mass spread of traditional deep learning ANN models. This may be resulting from less efficient training algorithms for deep SNNs, but also from other factors as the non-uniformity in encoding procedures of inputs  into discrete spike events. However, SNN have still been successfully used in many real-life applications including simulations in neurophysiology,  medical diagnostics,  pattern recognition (images, faces, handwriting, speech\dots), robotics control and much more. We refer the reader to recent survey works \cite{lobo2020spiking,yamazaki2022spiking} for more information.

\subsection{Architecture and function of SNPS}

The  idea of spiking neurons  was formalized mathematically by Ionescu, P{\u a}un, Yokomori in \cite{ionescu2006spiking}. The computing model introduced in \cite{ionescu2006spiking} is based on formal language and automata theory. Briefly, the model operates in discrete time steps and each neuron accumulates a discrete number of spikes sent to it by other neurons. Each neuron contains a set of action rules of two possible types: (i) a spiking rule which consumes a certain number $c$ of spikes and emits $p\le c$ spikes to other neurons; (ii) forgetting rule which just removes all spikes from the neuron. Each spiking rule is associated with a regular expression over integers, and it can be applied when the number of spikes present in the neuron matches this expression. Furthermore, its application can be delayed by a specific number of time steps. We refer the reader to the monograph \cite{10.5555/1738939} for a detailed formal definition.
	
In the past decade, many variants of SNPS models have been introduced. Some of these models are rather theoretical such as asynchronous SNPS \cite{cavaliere2009asynchronous}, SNPS with anti-spikes \cite{pan2009spiking}, SNPS with polarization \cite{wu2017spiking}, numerical SNPS \cite{wu2020numerical}, cell-like SNPS \cite{wu2016cell}, SNPS with rules on synapses \cite{song2014spiking}, SNPS with white-hole neurons \cite{song2016spiking}, SNPS with pre-computed resources \cite{ishdorj2010deterministic},  SNPS with communication on request \cite{wu2022tuning} etc. 
Furthermore, some variants of SNPS models contain features which are related to cell biology of neurons, such as SNPS with neuron division and budding \cite{pan2011spiking}, SNPS with structural plasticity \cite{cabarle2015spiking}, weighted SNPS \cite{wang2010spiking}, SNPS with astrocytes \cite{pan2012spiking}, SNPS with multiple channels \cite{peng2017spiking,liu2022parallel}, echo SNPS \cite{long2022echo}, SNPS with self-organization \cite{wang2016computational}, SNPS with scheduled synapses \cite{cabarle2017spiking}, gated nonlinear SNPS with applications in time series forecasting \cite{zhang2023prediction}, nonlinear SNPS with two outputs used for edge detection \cite{xian2023edge}, a fuzzy membrane control SNPS \cite{liu2023human} etc. The variants of the SNPS listed above mostly enrich the basic SNPS model with various additional features, on one hand. On the other hand, normal forms aimed at the study of restricted variants of the SNPS (while preserving their computational power) were studied in \cite{ibarra2007normal,garcia2007spiking}.

The formal framework of SNPS models has been experimentally verified in many practical applications ranging from control and optimisation to various artificial intelligence problems and finally to biological modelling problems.  The applications include areas such as fault diagnosis of power systems, pattern recognition, image processing, classification problems, image processing, time series forecasting and more. For more details we refer the reader to survey papers \cite{rong2018spiking, fan2020applications}.

\subsection{Comparison of SNN and SNPS models}  \label{section2}

Both SNN and SNPS belong to the category of third-generation (spiking) neural networks. 
The main objective of the SNN and SNPS models is the temporal encoding of information processed by the spikes of neurons. However, along with these similarities, the two models have also distinctive features.  We compare some of the properties of SNN and SNPS in Table \ref{tab:comp}. 

 \begin{table}[h]
\caption{Comparison of SNN and SNP models \label{tab:comp}}
{\begin{tabular}{lp{6cm}}
 
 \multicolumn{2}{c}{\bf Neuron activation condition}\\
 SNN& Membrane potential exceeds a threshold.\\
 SNPS& The number of accumulated spikes is in the set represented by a regular expression.\\\\

\multicolumn{2}{c}{\bf Inputs and outputs}\\
 SNN& Spike trains or real numbers encoded into/decoded from spike trains.\\
 SNPS& Mostly integers encoded as a number of spikes or discrete time delays. \\\\
 
\multicolumn{2}{c}{\bf Computation process}
\\
 SNN& Continuous rather than the discrete domain. Nuances in firing rules, no forgetting rules. \\
 SNPS& Operating in discrete time, simple spiking and forgetting rules. Colored spikes and anti-spikes to encode additional information. Fuzzy reasoning SNPS\cite{peng2013fuzzy}  models resemble SNN in operation.\\\\

\multicolumn{2}{c}{\bf Computability and Complexity}\\
 SNN& A few studies of Turing completeness\cite{potok2022neuromorphic,maass1996lower} and computational complexity\cite{fonseca2017using}. Neuromorphic completeness of some models has been discussed in \cite{zhang2020system,rhodes2020brain}. \\
 SNPS& Most variants are proved to be Turing complete, solving NP-compete or even PSPACE-complete problems\cite{leporati2007solving,pan2009spiking,song2014solving,cabarle2015spiking,sosik2019p,xu2013simple,gatti2022spiking}.\\\\

\multicolumn{2}{c}{\bf Training}\\
 SNN&A broad scope of methods - gradient descent-based, spike time dependent plasticity algorithms, substitution methods, approximate derivative methods, deep learning\dots\\
 SNPS & Due to the formal language / automata background, training even a single-layer model is rather difficult. Not so many results exist.\cite{chen2018computational,peng2010adaptive,wang2013adaptive,song2019spiking,chen2021survey,zhang2022layered,qiu2022deep,zhang2021dissolved,xue2023hypergraph,liu2022lstm,liu2023nonlinear,yang2023sddc,huang2023sentiment}. \\

\end{tabular}}
\end{table}

{ In summary, SNPS and SNN models are two independent third-generation neural network models based on communication using spikes. Main distinctive features of SNPS with SNN models are: 
(1) a formal language-theoretic structure and motivation of SNPS, in contrast with close biological inspiration of SNN;
(2)  SNN has different types of encoding schemes allowing for real-numbered inputs/outputs. In contrast, SNPS models encode information using the number of spikes or discrete time delays. There exist also discretized models of SNN with binary weights or even neuron activation\cite{lu2020exploring,kheradpisheh2020temporal} benefiting from temporal encoding of information. Therefore, they are still distinct from SNPS which operate in discrete time steps.
(3)  Much effort is devoted to study the computing capability SNPS models, most of which are Turing complete. The computing power of SNN models is less investigated. 
}

\section{Machine learning in SNN}  \label{section3}

Efficient machine learning of neural networks as SNN with discrete information processing is one of the most challenging problems in their research. As the forward pass of information through the network represents mostly a non-differentiable function, the classical gradient algorithms cannot be used and alternative solutions must be sought. The revolutionary progress of deep learning during the last decade accelerated this research and many types of machine learning and deep learning algorithms for SNN have been introduced. 

The weight modification process during learning in SNN is associated with timing of spikes between two connected neurons. The learning in SNN can be divided into three categories, i.e., (1) direct supervised learning (backpropagation-based etc.); (2) Unsupervised learning (spike timing-dependent plasticity -- STDP); often, weights on the synapses are modified using the information locally available. This learning process is called local learning and it resembles the learning process in the neurons present in a human brain. (3) Indirect supervised learning (conversion of an ANN to SNN). SNN obtained after translation of pre-trained ANN into an equivalent SNN can have higher representation power and capacity \cite{davidson2021comparison}. A brief summary of learning algorithms for SNN is given in Table \ref{tab02}.

\begin{table*}[tb!]
	\caption{Machine learning and deep learning algorithms based on SNN}
	\label{tab02}
	{
	\begin{tabular}{@{}p{6.6cm}p{1.8cm}p{7cm}@{} } 
  \hline
		Machine learning and deep    & Supervised/   & Real-life     \\
  learning models based on SNN &  Unsupervised     &  Application  \\
         
  \hline
		SpikeProp with a BP learning rule\cite{bohte2000spikeprop} / \par with Levenberg-Marquardt algorithm \cite{silva2005application} / \par with multi-spike neurons \cite{booij2005gradient} & Supervised  & Non-linear classification tasks /\par  classification of Poisson spike trains \par \mbox{ }\\

  \hline
		SpikeProp with QuickProp or Rprop\cite{xin2001supervised,mckennoch2006fast} & Supervised  & XOR and Fisher Iris data sets
\\
    \hline 
		Multi-SpikeProp\cite{ghosh2009new} & Supervised  & Epilepsy and seizure detection 
\\
    \hline
	
		Back-propagation with momentum\cite{fu2017improving} & Supervised  & Wisconsin breast cancer classification 
\\
     \hline
		Spatio-temporal back propagation\cite{wu2018spatio} & Supervised  &  MNIST / N-MNIST dataset  \\

      \hline
		SuperSpike + Hebbian three-factor rule\cite{zenke2018superspike,zenke2021remarkable} & Supervised  & Randman, MNIST, SHD, RawHD \\

      \hline
		ReSuMe (Remote Supervised Method)\cite{ponulak2010supervised} & Supervised  & Learning spike trains \\

      \hline
		Chronotron with STDP\cite{florian2012chronotron} & Supervised  & Precise spike trains  \\

      \hline
		BP-STDP for multi-layer SNNs\cite{tavanaei2019bp} & Supervised  & XOR, Iris data, MNIST dataset  \\

   \hline
		Supervised STDP (SSTDP) \cite{liu2021sstdp} & Supervised  & CALTECH 101 / MNIST / CIFAR-10 datasets \\

      \hline
		Symmetric STDP\cite{hao2020biologically} & Supervised  & Fashion-MNIST dataset  \\

     \hline


%
 
    SPAN (spike pattern association neuron)\cite{mohemmed2012span} & Supervised  & Spike pattern classification  \\

    \hline
	Spike train kernel learning (STKLR)\cite{lin2016new,ma2018supervised} & Supervised  & LabelMe image dataset\\

    \hline
		STDP variants\cite{vigneron2020critical} & Unsupervised  & Pattern recognition  \\

    \hline
		Locally-connected SNN with STDP\cite{saunders2019locally} & Unsupervised  & Learning image features  \\

    \hline
		Spiking CNN with STDP\cite{tavanaei2017multi} & Unsupervised  & MNIST digit dataset   \\

     \hline
		Self-organizing SNN\cite{hazan2018unsupervised}  & Unsupervised  & Decision-making system  \\

     \hline
		SpikeDyn\cite{putra2021spikedyn}  & Unsupervised  & Image classification  \\

     \hline
		SpiCNN (deep spiking CNN)\cite{lee2018deep} & Supervised  & CALTECH / MNIST  datasets \\

     \hline
		Deep spiking CNN using Tensorflow\cite{vaila2019feature} & Supervised  & MNIST and NM-NIST datasets  \\

     \hline
		Deep residual learning SNN\cite{fang2021deep} & Supervised  &  ImageNet, DVS Gesture, CIFAR10-DVS \\

      \hline
		Spiking-Yolo\cite{kim2020spiking} & Supervised  & Object detection \\

      \hline
		Spiking recurrent NN\cite{kim2019simple} & Supervised  & Cognitive tasks \\

     \hline
		Adaptive spiking RNN\cite{yin2021accurate} & Supervised  & Speech and gesture recognition \\

      \hline
		Spiking convolutional RNN\cite{xing2020new} & Supervised  & Hand gesture recognition \\

     \hline
 
		Spiking deep belief network (DBN)\cite{o2013real} & Supervised  & MNIST handwritten digits \\

     \hline
 
		Spiking DBN on SpiNNaker\cite{stromatias2015scalable,stromatias2015live} & Supervised  & MNIST handwritten digits  \\

     \hline
 
		Spiking DBN with Siegert neuron model\cite{fatahi2016towards} & Supervised  & Face recognition
  \\

   \hline
     
        Spiking DBN\cite{ganapathy2020emotion} & Supervised  & Emotion analysis from electrodermal signals \\

     \hline
 \end{tabular}}
	
\end{table*}

\subsection{Supervised learning algorithms for SNN}

Supervised learning algorithms for SNN can be classified according to their principle into three basic groups: gradient descent-based, synaptic plasticity-based and spike train convolution-based. We provide a few examples for each group. 

\subsubsection{Gradient descent-based algorithms}

The \textit{SpikeProp} \cite{bohte2000spikeprop} training algorithm was introduced in 2002 by Bohte et al. It is a feedforward multilayer SNN framework and state-of-the-art error-backpropagation learning rule for SNN. SpikeProp has been a subject to many extensions and improvements. An improved variant of the SpikeProp algorithm was introduced by Silva et al. in  \cite{silva2005application} using the Levenberg-Marquardt backpropagation method of training. Another extension of the SpikeProp to SNN with neurons that emit multiple spikes was introduced in 2005 by Booji and Nguyen\cite{booij2005gradient};

Resilient propagation (RProp) \cite{mckennoch2006fast} was designed to speed up the training process with a learning rate adaptation technique depending on the sign of the gradient. Also the QuickProp principle from analog ANNs has been used in \cite{xin2001supervised,mckennoch2006fast} and others to speed up the training process  of SpikeProp, thanks to useful assumptions about the data and error surface. One of the main features of  Quickprop algorithms is the use of Newton's method for the purpose of minimization of one-dimensional function.

Multi-SpikeProp was introduced in 2009 by Ghosh-Dastidar and Adeli \cite{ghosh2009new}. It can be seen as an extension of the SpikeProp to train MuSpiNNs (Multi-Spiking Neural Networks) which led to a dramatic increase of efficiency of the model (the authors claim the improvement by two orders of magnitude).

Back-propagation with momentum for SNN was presented in \cite{fu2017improving}. Learning performance of gradient descent algorithms was improved by a combination of back-propagation with momentum, QuickProp and heuristic rules. 

\textit{Spatio-temporal back propagation} introduced in \cite{wu2018spatio} by Wu et al. is a combination of layer-by-layer spatial domain phase and time-dependent temporal domain phase.

\subsubsection{Synaptic plasticity-based algorithms}

 These algorithms are mostly based on the spike time-dependent plasticity (STDP) learning mechanism \cite{caporale2008spike}, applicable both in supervised and unsupervised training schemes. 	The STDP learning rules are inspired by Hebbian learning. Synaptic strengths are adjusted based on the correlation of the pre-synaptic and post-synaptic firing times.

 The \textit{ReSuMe (Remote Supervised Method)} \cite{ponulak2010supervised} was introduced in 2010 by Ponulak and Kasi\'nski. The algorithm is based on a combination of STDP and anti-STDP and later it provided a base for several imporoved variants. The SNN trained in ReSuMe are capable of learning and reproducing complex patterns of spikes.

\textit{Chronotron} introduced in \cite{florian2012chronotron} is  an algorithm based on information encoding in the form of precise timing of a train of spikes. Neurons are trained to generate a desired  output spike train in a response to a specific spike patterns received as input. Two learning rules have been introduced: E-learning with high memory capacity, and more biologically plausible I-learning.

More recently, \textit{BP-STDP alorithm} \cite{tavanaei2019bp} has been introduced in 2019, using the concepts of gradient descent and STDP applicable in multi-layer SNN, potentially also in deep SNN .

Among other recent algorithms combining gradient descent and STDP we can name Supervised STDP \cite{liu2021sstdp}. Further variants of supervised learning algorithms based on STDP have been used, for instance, for training of multi-layer photonic SNN \cite{xiang2020training}. Hao et al.\cite{hao2020biologically} proposed a neuroscience-plausible  symmetric STDP rule which provided competitive results in classification tasks.
From many other applications of STDP-based learning we mention, e.g., touch modality classification  \cite{dabbous2021touch}, or MNIST image dataset learning \cite{chang2018interchangeable,hu2017stdp}. 

\subsubsection{Spike train convolution algorithms}
\textit{SPAN (spike pattern association neuron)} \cite{mohemmed2012span} stands for a learning algorithm based on the Widrow–Hoff rule and a transformation of spike trains into analog signals via an $\alpha$-kernel function. 

The use of kernel function turned out to be a promising approach in this class of learning methods. A spike train
kernel learning rule (STKLR) with various kernel functions has been proposed in \cite{lin2016new}. The algorithm was later extended from single-layer to multi-layer networks and several applications have been reported, see, e.g., \cite{ma2018supervised}.

\subsubsection{Resume}

In 2020, Wang et al.\cite{wang2020supervised} prepared a comprehensive survey on supervised learning algorithms for SNN, including experimental comparison of a selected group of algorithms. In general it can be concluded that for feedforward SNNs, algorithms based on STKLR\cite{lin2016new,lin2017supervised} performed very well and they gave the best score in several tests. However, due to the diverse nature of learning approaches and test criteria, it is rather difficult to provide a unified scoreboard \cite{wang2020supervised}. Further supervised learning methods for multilayer SNN can be found, e.g., in \cite{sporea2013supervised,zenke2018superspike,luo2022supervised,taherkhani2018supervised,xie2016efficient,lin2021supervised} and many more. A recent study on supervised training algorithms for SNN inspired by deep learning methods has been published in \cite{eshraghian2023training}.

\subsection{Unsupervised learning algorithms for SNN}  
Unsupervised learning models deal with unlabelled data and, instead of supervision, they identify hidden patterns inside the data on their own. A vast majority of unsupervised learning algorithms in SNN are based on the concept of synaptic plasticity. Indeed, the concept of spike time-dependent plasticity (STDP) introduced in the previous section forms a base of many SNN training algorithms. STDP rules can have many variants such as $C_a$-STDP, $C_m$-STDP, M-STDP, P-STDP etc. A summary of these and further STDP variants can be found in \cite{vigneron2020critical}. The use of STDP is mostly found in unsupervised learning \cite{vigneron2020critical}, although it is also applied in many supervised and reinforcement learning algorithms, often in combination with gradient descent. In this section we provide several examples of successful unsupervised learning algorithms for SNN.

 In Saunders et al. \cite{saunders2019locally}, an unsupervised feature learning for locally connected spiking neural networks was conceptualized. The paper introduced locally connected SNN learning features of images with locally connected layers of SNN. The model uses the STDP learning algorithms providing, thanks to the local connections, faster convergence than original SNN. 

Tavanaeri and Maida \cite{tavanaei2017multi} proposed a spiking CNN model containing spiking convolutional-pooling layer and feature discovery layer. An unsupervised local learning has been used to train kernels of the convolution layers and the feature discovery layer consists of probabilistic neurons. A probabilistic STDP rule has been used for training convolution layers of the spiking CNN. The network performed competitively on the MNIST digit dataset. 

Unsupervised learning of self-organizing spiking neural networks has been introduced in \cite{hazan2018unsupervised} and applied to train  a rapid decision-making system. The architecture is inspired by classical Kohonen's self-organizing maps with the capability of clustering unlabeled datasets in an unsupervised manner.
	
\textit{SpikeDyn} \cite{putra2021spikedyn} is the name of a framework based on SNN with unsupervised learning in dynamic environments was introduced in. Its design focuses on a restriction of neuronal operations and memory and energy consumption, leading to the decrease of training energy consumption by approx. 50\%. 
	
In  \cite{weidel2021unsupervised}, a combined framework of  unsupervised learning algorithm and clustered connectivity has been introduced, confirming experimentally an enhanced reinforcement learning capability in SNN.

	
		

		

\subsection{Deep SNN} 

Deep neural network (DNN) architectures are primarily based on real-valued ANN and they are composed  of multiple hidden layers between the input and the output neurons. DNN are capable of learning complex patterns present inside the data. In the last decade, DNN have become a useful tool for solving problems in image processing, natural language processing, pattern recognition, self-driving cars etc. However, training of deep spiking neural networks is still a very challenging problem. As we have already mentioned,  the backpropagation algorithm cannot be directly implemented for deep SNN due to their non-differentiable nature. A review of supervised and unsupervised learning methods for training of deep SNN can be found in \cite{tavanaei2019deep}. In this section, we briefly discuss main features of several deep SNN models and their real-life applications such as spiking CNN, spiking RNN, spiking DBN etc. 
For instance, the performance of the SNN in solving pattern recognition tasks has been very promising \cite{tavanaei2019deep}. A recent study\cite{eshraghian2023training} summarizes deep learning methods and approaches applicable in learning of deep SNNs.

\subsubsection{Spiking convolutional neural networks}

Spiking CNNs are mainly used for image processing and two-dimensional grids, object detection etc. They are mostly based on conversion of already trained CNN into a spiking CNN which is more efficient and consume less energy.
In \cite{lee2018deep}, Lee et. al introduced a deep spiking neural network with a hierarchy of stacked convolution layers, i.e., SpiCNN (deep Spiking Convolutional Neural Network). The convolution kernels are trained using spike time-dependent plasticity (STDP) rules which is a well-known method in unsupervised learning.  

A study on the implementation of spiking CNNs using Tensorflow is performed in \cite{vaila2019feature}. A spike-element-wise (SEW) ResNet with residual learning in a deep SNN was presented in \cite{fang2021deep}. The  authors claim primacy in a direct training of a deep SNNs with more than 100 layers. A  channel-wise normalization and neurons with imbalanced threshold were used in \cite{kim2020spiking} to implement a spiking version of the Yolo network, performing on the level comparable with Yolo Tiny, but with energy consumption lower by two orders of magnitude.

\subsubsection{Spiking recurrent neural networks} 
Recurrent neural networks (RNNs) with classical units as the LSTM (Long short-term memory) or GRU (gated recurrent unit) are primarily used to process sequential data. In this section we focus on some recent implementations of RNNS based on spiking neurons.

In  Kim et. al.\cite{kim2019simple}, a  biologically realistic framework for spiking RNNs is introduced and shown to be capable of performing many elementary cognitive tasks. 
In \cite{yin2021accurate}, a novel spiking RNN model with adaptive multiple-timescale spiking neurons was introduced by Yin et al. The model reaches the performance of analog RNNs with the energy consumption lower by one to three orders of magnitude. 
Xing et al.\cite{xing2020new} conceptualized a new variant of SNN which combines the properties of convolution and recurrent neural networks. The model is called spiking convolutional recurrent neural network (SCRNN) and it uses a supervised training algorithm called SLAYER (Spike Layer Error Reassignment). 
A temporal spiking recurrent network, another novel variant of spiking RNN, was introduced in \cite{wang2019temporal}. The model has been applied for video recognition and its  performance is competitive against SOTA CNN-RNN models in the literature. 
	
\subsubsection{Spiking deep belief networks} 

 Deep belief network (DBN) \cite{bengio2009learning} is a generative neural network model using stacked layers of Restricted Boltzmann Machines (RBMs).  It is used mostly to reconstruct or classify input data (e.g., images ).
  In \cite{o2013real}, O'Connor et al. proposed a spiking deep belief network. These models can efficiently classify MNSIT handwritten digits. Later studies in \cite{stromatias2015scalable,stromatias2015live} showed that DBN on SpiNNaker can achieve classification performance of 95$\%$ on the same dataset.
 
In \cite{xue2017real}, Xue et al. investigated the object recognition capacity of spiking DBN models. 
Face recognition capability of spiking DBN models has been evaluated on ORL dataset in \cite{fatahi2016towards}. Spiking DBN models have also been used for emotional analysis using electrodermal signals \cite{ganapathy2020emotion}. The study on \textit{spiking convolution deep belief networks} has gained interest of researchers due to its capacity to learn high-level features from dynamic vision sensor datasets \cite{kaiser2017spiking}. However, the accuracy of the spiking DBN on the MNIST dataset was shown lower than that of other types on SNNs\cite{tavanaei2019deep}.

\section{Learning algorithms for SNPS}  \label{section4}

\begin{table*}[htb]
	\caption{Machine learning and deep learning algorithms based on SNPS}
	\label{tab03}
	
	\begin{tabular}{@{}p{6.8cm}p{3cm}p{5,2cm}@{} } 
   \hline
		Machine learning and deep    & Supervised /   & Real-life     \\
  learning models based on SNPS &  Unsupervised     &  Application  \\
	
   \hline
		SNPS with Hebbian learning \cite{chen2018computational} & Unsupervised  &  Identification of  nuclear export signal 
  \\
  \hline
		Adaptive SNPS with 	 Widrow-Hoff learning algorithm  \cite{peng2010adaptive}  & Supervised  &  Fault diagnosis 
  \\
   \hline
		Adaptive  fuzzy SNPS with  Widrow-Hoff learning algorithm \cite{wang2013adaptive}  &  Supervised     &  Fault diagnosis  
  \\
   \hline
		SNPS with learning  function  \cite{song2019spiking}  (Hebbian) &   Unsupervised      &   Recognize digital English letters  \\
 \hline
		Associative memory network based on SNPS with white holes
		 + Hebbian learning  \cite{chen2021survey}  	&   Unsupervised      &  Identification of   digits  
  \\
    \hline
		Layered SNPS \cite{zhang2022layered} &   Supervised      &    Classification problem 
  \\
   \hline
  Deep dynamic SNPS (SNPS + CNN) \cite{qiu2022deep} &  Supervised (ensemble learning)     &   Organ segmentation  
  \\
 \hline
		Hypergraph-based SNPS \cite{xue2023hypergraph} &   Supervised  (ensemble learning)     &   predicting the survival time of glioblastoma  patients 
\\
  \hline
		LSTM-SNP \cite{liu2022lstm} & Supervised         &   Time series forecasting 
\\
   \hline
		Nonlinear SNPS \cite{liu2023nonlinear} &   Supervised  (linear machine learning)      &      multivariate time  series forecasting 
\\
   \hline
		Spiking neural P-type Dual-channel  dilated 	convolutional network (SDDC-Net) \cite{yang2023sddc}  &  Supervised       &  Retinal vessel   segmentation 
\\
   \hline
		(bidirectional) LSTM-SNP \cite{huang2023sentiment,liu2023attention}  &   Supervised       &  Sentiment classification 
\\
   \hline
		Load forecasting non-linear SNPS (LF-NSNPS) \cite{long2022time} &   Supervised       &   Electricity load forecasting     \\
   \hline
Learning SNPS with belief  AdaBoost  \cite{zhang2021dissolved} &   Supervised learning      &   
Fault diagnosis   
\\
   \hline
	\end{tabular}
\end{table*}

In recent years, machine learning and deep learning frameworks have been introduced also for SNPS models. In \cite{gutierrez2009hebbian}, Guti{\' e}rrez-Naranjo et al. introduced the first study of introducing Hebbian learning in the SNPS framework. This study was further continued by many authors and the SNPS model with Hebbian learning method has been applied to various tasks. Among them we mention, e.g., the analysis of nuclear export signals \cite{chen2018computational}, recognition of digital English letters \cite{song2019spiking}, identification of digits \cite{chen2021survey}. Hebbian learning belongs to the category of unsupervised machine learning.
The model introduced in \cite{chen2018computational} has two modules, i.e., (i) input and (ii) predict module. The rules used in this model are $E / a^{c} \rightarrow a^{p}$ and $  a^{s} \rightarrow \lambda$.  The Hebbian learning strategy is used in the predict module. It is important to note that the topology of the model remains unchanged in the input module. However, it changes in the predict module. To perform the task of identification of nuclear export signal (NES), at first, it is encoded into a binary string. Moreover, the model in \cite{chen2018computational} has a better precision rate than NES-REBS, Wregex, ELM and NetNES. In \cite{song2019spiking}, Song et. al. introduced an SNPS-based machine-learning model for recognition of digital English letters. The spiking rule used in \cite{song2019spiking} is $E / a^{c} \rightarrow a$. Similarly, as in \cite{chen2018computational}, synaptic connections change dynamically using learning function. Also,  this model is divided into two modules, i.e.,   in (i) the input module; (ii) the recognize module. The recognize module is divided into innermost, middle and outermost layers. This model performs better than SNNs in recognizing letters with noise. In \cite{chen2021survey}, Hebbian learning is used for the identification of digits. The working of the model is inspired by the Hopfield network. It uses only spiking rules of the form $a^{*} / a^{all} \rightarrow a$ and the set of input and output neurons are the same. Finally, the synaptic weights are updated using the energy function.

Over the years, researchers from the membrane computing community have developed several supervised machine learning algorithms based on spiking neural P systems. The first of them, based on the Widrow-Hoff rule, was proposed in 2010 by Peng and Wang \cite{peng2010adaptive}. In this model, a neuron $\sigma_i = (\alpha_i, \omega_i, R_i); $ has rules of the form $ E / a^{\alpha} \rightarrow a^{\alpha}$ where $\alpha$ represents the potential of the neuron $\sigma_i$ instead of number of spikes.  
This study was further extended in \cite{wang2013adaptive}. In \cite{wang2013adaptive}, Wang et. al. introduced adaptive fuzzy spiking neural P systems. This model is capable of performing dynamic fuzzy reasoning using the weighted fuzzy reasoning rules, where the fuzzy reasoning algorithm and the training mechanism are based on the firing mechanism of the neurons. 
Moreover, this model has two types of neurons, i.e.,  proposition neuron and rule neuron. A proposition neuron is a $4$-tuple  $\sigma_i = (\alpha_i, \omega_i, \lambda_i, r_i$), where $\alpha_i \in [0,1], \omega_i$ (weight vector), $\lambda_i \in [0,1)$ (threshold) and has spiking rule $E/ a^{\alpha} \rightarrow a^{\alpha}$. Similarly, rule neuron is represented by a $4$-tuple $\sigma_{ri} = (\alpha_i, \gamma_i, \tau_i, r_i)$, where $\alpha_i, \gamma_i \in [0,1]$ and has spiking rule $ E / a^{\alpha} \rightarrow a^{\beta}, \alpha, \beta \in [0,1]$. Moreover, synaptic connections exist only between proposition and rule neurons. 
In \cite{zhang2022layered}, a new variant of SNPS model, i.e., layered spiking neural P system was introduced by Zhang et. al to solve classification problems. This model contains multiple weighted SNPS and weight-adjusting rules. Furthermore, the performance of layered SNPS in solving classification problems has been verified on the MNIST dataset. 
  The model in \cite{zhang2022layered} is highly-flexible, simple and has a
  multi-layer weighted SNPS with adaptive weight adjustment rule. Similarly, as in \cite{wang2013adaptive,peng2010adaptive}, the model contains proposition neurons and rule neurons with similar spiking rules. However unlike in \cite{wang2013adaptive,peng2010adaptive}, it contains a high dimensional encoder (which exists between the pre-processing and input layer) and a weighted fuzzy SNPS classifier. It has input, hidden, comparison, selection and output layers. The main difference between WFSNPS and LSNPS is that the first one has one layer and the latter has more than one layer. Furthermore, it has excellent learning ability and convergence.
  
  In \cite{zhang2021dissolved}, Zhang et. al introduced a novel learning spiking neural P system with the belief AdaBoost for identification of faults in oil-immersed power transformers. Similarly, as in \cite{wang2013adaptive,peng2010adaptive,zhang2022layered}, it has proposition and rule neurons. 
Moreover, the experimental implementation of the models shows that it has higher accuracy in the identification of faults in comparison with back-propagation neural networks, support vector machine etc.

In \cite{xue2023hypergraph}, a new variant of hypergraph-based neural network, i.e., hypergraph-based numerical P system was introduced. In recent years, hypergraph neural networks have become an attractive direction for research due to their superior performance in many domains.  The P system variant of the hypergraph neural network in \cite{xue2023hypergraph} can represent the high-order neural connections and can be implemented for the segmentation of tumours/organs in medical images. This model also outperforms state-of-the-art methods which are based on hippocampal datasets and multiple brain metastases datasets. 
It consists of three new classes of neurons with Plane, Transmembrane and Hierarchical rules to describe higher-order relationships among neurons. The time consumed for the segmentation of images was also reduced. 
In \cite{qiu2022deep}, Qiu et. al. proposed a novel organ segmentation method based on a deep dynamic SNPS model. It is a combination of features from SNPS and CNN. It has two subsystems (i) locating pancreas (ii) accurate segmentation of the pancreas. Accuracy of this model is lower than some state-of-the-art methods. Again, the system is represented as $\Pi = (O, C, \sigma_{iL}, syn, M, \sigma_{in}, \sigma_{out})$ where $C$ represents coefficients of the spike and the neuron   $\sigma_i = (s_i, r_i)$ has rules of the form $[a^{s} \rightarrow a^{p}]_i$, $[a^{s} \rightarrow \lambda]_i, [a^{s}]_i \rightarrow [\lambda]_j [\lambda]_k (i \neq j \neq k)$.

Long short-term memory is a novel variant of neural networks. Based on the properties of LSTM, a new variant, i.e., LSTM-SNP model was introduced in \cite{liu2022lstm}. LSTM-SNP also has been implemented for time series forecasting. 
It is a recurrent-type model and can process sequential data. Again, it  has
nonlinear spiking mechanisms (nonlinear spike consumption and
generation) and nonlinear gate functions (reset, consumption,
and generation gates). The neurons in this model have spiking rule of the form $T / g(x_i) \rightarrow f(x_i)$ where $T$ represents a threshold,  $g(x_i)$  is a linear or non-linear function, $f(x_i)$ is a non-linear function.  LSTM-SNP model has better predictive performance than some state-of-the-art prediction models.
Also, it has a lower predictive performance for some time series. 

In \cite{liu2023nonlinear}, Liu et. al proposed a new variant of SNPS model as well as a recurrent neural network, i.e., SNP systems with autapses. These models contain three non-linear gate functions and a recurrent-type prediction model is constructed based on this model. 
In summary, it has four mechanisms, i.e.,  (1) Non-linear spiking; (2) recurrent mechanism; (3) Non-linear gate mechanism; (4) autapses mechanism.
Furthermore, this model has been used in chaotic time series and experimental results demonstrate the effectiveness of this model in time series forecasting. 
   The neurons in \cite{liu2023nonlinear} are of the form $\sigma_i = (u_i, r_i, (\rho_1^{i}, \rho_2^{i}, \rho_3^{i})$ where  $u_i$ represents internal state and the gate function is represented by $\rho^i$. The spiking rule used in this model is of the form  $T / a^{g(u)} \rightarrow a^{f(u)})$ where $T$ is a threshold and $f,g$ are non-linear functions. This model also can Predict  Lorenz
chaotic time series efficiently.
In \cite{long2022time}, Long et. al. used non-linear spiking neural P systems for MST-based time series forecasting. In this method, wavelet transformations are used for the transformation of time series data into frequency domain data during the training phase. Next, the non-linear SNPS model is trained using the frequency domain data. This model is also capable of generating sequential data automatically as prediction results. More specifically, the proposed method in \cite{long2022time} is capable of processing non-stationary time series problems. The spiking rules used in \cite{long2022time} is similar to the rules in \cite{liu2023nonlinear}. However, the model has some extra properties such as (i) synaptic weights are associated with the neurons; (ii) neurons do not have any forgetting rules and have only one spiking rule. Moreover, the constraint $g(x) \geq f(x) \geq 0$ has been removed from the spiking rules.

In \cite{huang2023sentiment}, a modified LSTM-SNP model, i.e., bidirectional LSTM-SNP (BiLSTM-SNP) was introduced by Huang et. al. Moreover, aspect-level sentiment classification model based on BiLSTM-SNP has been introduced. The semantic correlation between the aspect word and content word has been investigated using this model and to understand the effectiveness of the model, it has been implemented on English and Chinese datasets. 
This model is a variant of non-linear SNPS and it is a combination of a forward LSTM-SNP (which extracts the contextual semantic information towards aspect word) and a backward LSTM-SNP (which obtains the semantic correlation between aspect word and content words). It also outperforms some variants of LSTM models.

Recently, convolution neural networks have gained prominence due to their image and video processing capacity and perform tasks such as facial recognition, object detection, image classification etc very efficiently. 
In \cite{yang2023sddc}, 
an U-shaped deep CNN model, i.e., spiking neural P-type Dual-channel dilated convolution network (SDDC-Net) is proposed by Yang et. al. 
It has the classic encoder-decoder architecture containing convolutional neurons which makes it different from the standard SNPS model. 
This model has excellent segmentation, and generalization ability and SNP U-Net increases the accuracy.

\section{Conclusion} \label{section6}

This survey focuses on the relation of spiking neural networks (SNN) and their variant popular in the field of membrane computing, i.e, spiking neural P systems (SNPS). We compared these models based on their properties, architectures and real-life applications. Most importantly, we discussed a range of machine learning algorithms for SNN and SNPS. In order to keep this study to a reasonable length, we could not cover all related topics, such as the issue of real-time recurrent learning (RTRL), for which we refer to the recent source \cite{eshraghian2023training}.

Although the research in both SNN and SNPS provides a series of theoretical results and interesting applications, it is fair to note that their efficient learning remains a challenging problem and their applicability in deep learning tasks cannot recently compete with that of real-valued  deep ANN models. Therefore, among open and challenging research problems in SNN and SNPS fields we mention first an efficient training of multi-layer SNN and SNPS. An interesting topic is their possible hardware implementation which promises a lower energy consumption than in the case of classical ANN. Thanks to their closer relation to biological brain structure, a possible use of SNN / SNPS models in sentiment classification seems interesting. Another promising application topic is the analysis of time series where the potential of spiking neuronal model is high due to their dynamical representation of data patterns \cite{ghosh2009spiking}. Finally, an efficient implementation of the spike timing-dependent plasticity learning \cite{vigneron2020critical} in SNPS could greatly improve their learning potential.

\bmhead{Acknowledgements}
Supported by the Silesian University in Opava under the Student Funding Plan, project SGS/11/2023.

\bibliography{SNN_SNP}

\end{document}